%% file: main.tex
\documentclass[10pt,twocolumn,letterpaper]{article}

\usepackage{cvpr}              
\usepackage{multirow} 
\usepackage{graphicx}
\usepackage{tablefootnote}
\usepackage{wrapfig}
\usepackage{amsmath}
\usepackage{adjustbox}
\usepackage{arydshln}
\usepackage[table]{xcolor} 
\usepackage{colortbl}
\usepackage{amssymb}

\input{preamble}
\definecolor{cvprblue}{rgb}{0.21,0.49,0.74}
\usepackage[pagebackref,breaklinks,colorlinks,allcolors=cvprblue]{hyperref}

\def\paperID{16008} 
\def\confName{CVPR}
\def\confYear{2026}

\title{ControlThinker: Unveiling Latent Semantics for Controllable Image Generation \\
through Visual Reasoning}

\author{%
    \fontsize{12}{14}\selectfont
    \textbf{Feng Han}\textsuperscript{1,2*},
	\textbf{Yang Jiao}\textsuperscript{1,2*},
	\textbf{Shaoxiang Chen}\textsuperscript{3},
    \textbf{Junhao Xu}\textsuperscript{1,2},
    \textbf{Jingjing Chen}\textsuperscript{1,2}, \\
	\textbf{Yu-Gang Jiang}\textsuperscript{1,2} \\
    \textsuperscript{1}Shanghai Key Lab of Intell. Info. Processing, School of CS, Fudan University\\
    \textsuperscript{2}Shanghai Collaborative Innovation Center on Intelligent Visual Computing\\
    \textsuperscript{3}MiniMax \\
}

\begin{document}


\maketitle
\renewcommand{\thefootnote}{}
\footnotetext{$^{*}$Equal contribution.}
\renewcommand{\thefootnote}{\arabic{footnote}}

\input{sec/0_abstract}    
\input{sec/1_intro}
\input{sec/2_related_work}

\input{sec/3_method}

\input{sec/4_experiment}
\input{sec/5_conclusion}
{
    \small
    \bibliographystyle{ieeenat_fullname}
    \bibliography{main}
}

\input{sec/X_suppl}

\end{document}

%% file: sec/0_abstract.tex
\begin{abstract}
The field of controllable image generation has seen significant advancements, with various architectures improving generation layout consistency to control signals. However, contemporary methods still face challenges in bridging the semantic gap between input text prompts with sparse semantics and the target images, often producing images that are misaligned with the prompt and exhibit artifacts. To address this challenge, we propose ControlThinker, a novel framework that employs a "comprehend-then-generate" paradigm. Firstly, by incentivizing the visual reasoning capability of a MLLM, latent semantics from control images are mined to enrich text prompts. This semantic-enriched text prompt then seamlessly aids in image generation without the need for additional complex modifications. Furthermore, to tackle the uncertainty arising from the ambiguity of control images, we encourage broader exploration of reasoning trajectories and select the optimal one using a metric-based output reward model (ORM). Extensive experimental results demonstrate that ControlThinker effectively mitigates the semantic gap between raw text prompts and target images, resulting in improved visual quality and semantic consistency across a wide range of benchmarks.


\end{abstract}

%% file: sec/1_intro.tex
\section{Introduction}
\label{sec:intro}

\begin{figure}[t] 
  \centering 
  \includegraphics[width=\linewidth]{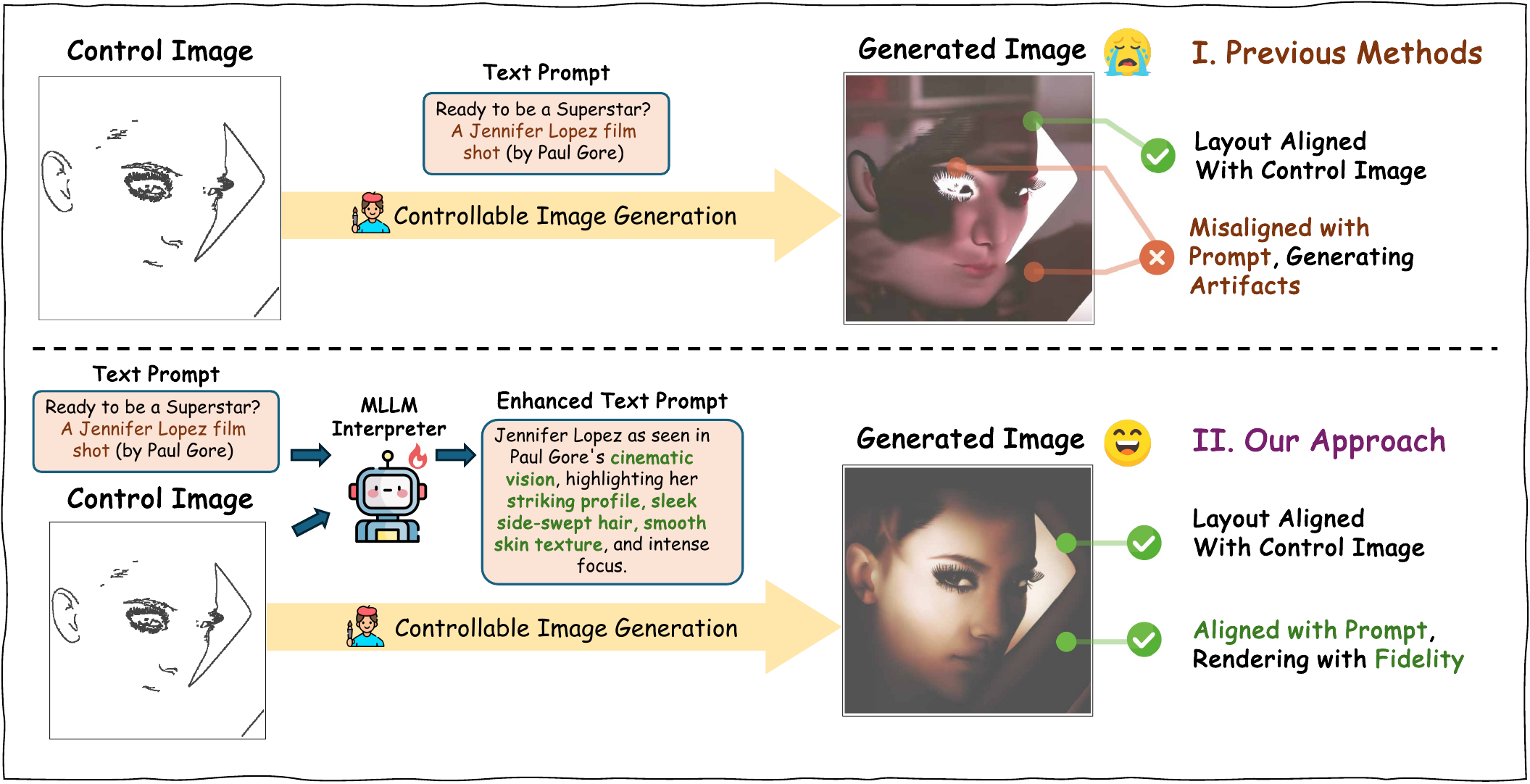} 
  \caption{Comparison between our ControlThinker and previous pipelines. Unlike previous methods that struggle to generate images adhering to the text prompt, ControlThinker leverages an MLLM as an input comprehender to enrich text semantics, enabling more accurate generation.} 
  \label{fig:figure1} 
  \vspace{-5mm}
\end{figure}

In recent years, significant advancements~\cite{rombach2022high,sun2024autoregressive,flux2024,chen2025janus,jiao2025unitoken} have been made in text-to-image generation, marked by enhanced comprehension of diverse textual prompts and improved aesthetic refinement in synthesized images. To achieve finer-grained control over image details, another line of research~\cite{zhang2023adding,li2024controlnet++,li2024controlar,qin2023unicontrol} incorporates images depicting low-level scene properties as conditional inputs. 
The prevailing paradigm in this field builds upon pre-trained text-to-image foundation models, augmenting them with specialized modules to enhance visual controllability \citep{sun2024anycontrol}.
Within this domain, two major research strands are commonly explored. The first, which has been a longstanding focus, emphasizes diffusion-based architectures to enhance the layout accuracy of generated images. More recently, a second strand has emerged, employing auto-regressive generation architectures and concentrating on adapting visual controls to these models.

\begin{figure*}[t] 
  \centering 
  \includegraphics[width=0.8\textwidth]{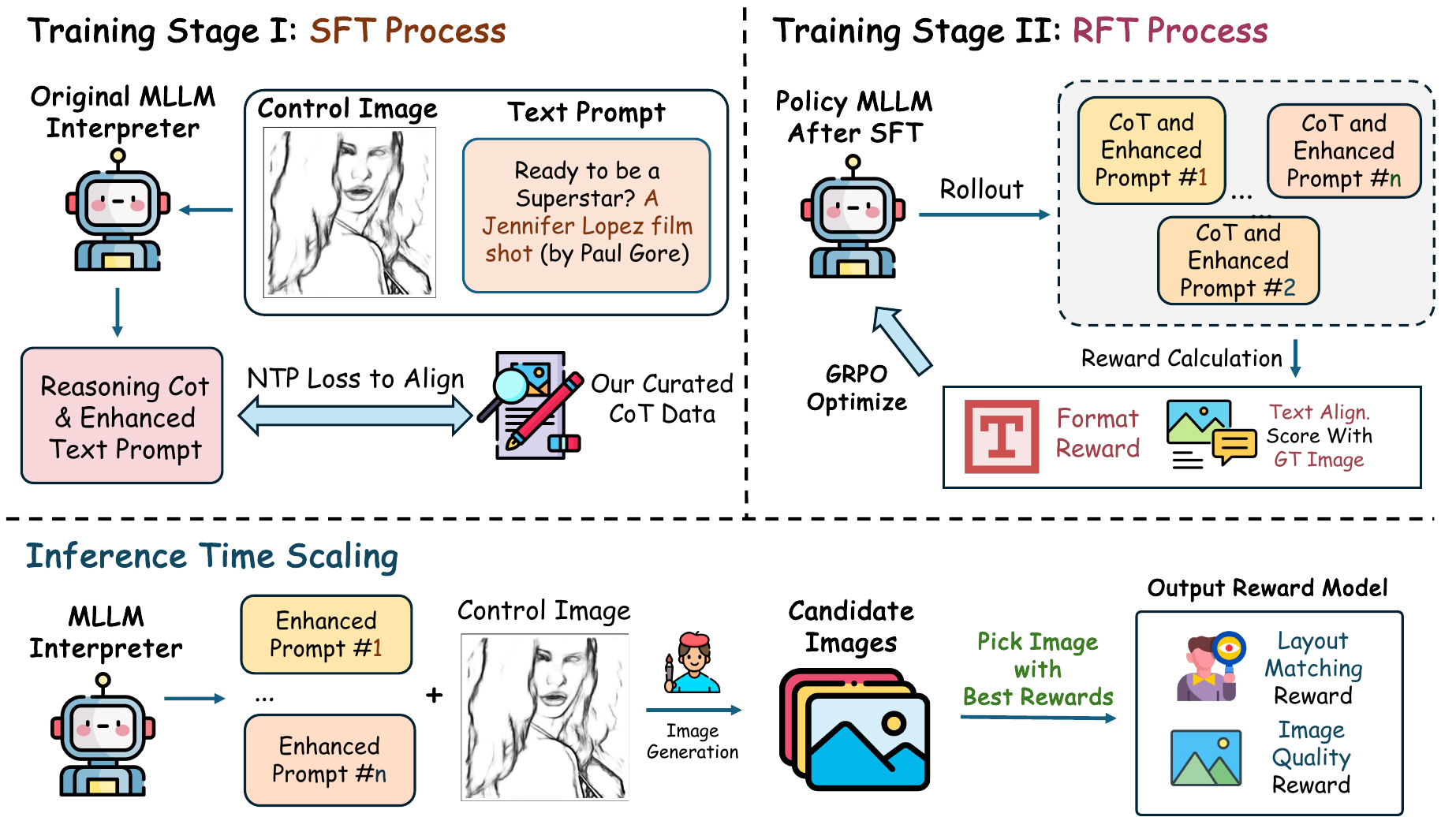} 
  \caption{The SFT and RFT two stage training paradigm and Inference Time Scaling method of our ControlThinker.} 
  \label{fig:figure2} 
  \vspace{-5mm}
\end{figure*}

Despite considerable advancements, current methods largely neglect the substantial semantic gap between text prompts and target images, constraining the full potential of pre-trained controllable image generators. As shown in Fig.~\ref{fig:figure1}, the ambiguity of input text prompts\footnote[1]{This is an inherent limitation of text prompts in existing controllable image generation datasets, as evidenced by the statistics presented in Fig.~\ref{fig:word-count}.} fails to provide semantically dense guidance for pixel-wise image synthesis. This compels the model to construct high-fidelity image content solely from low-level layout semantics of control images (e.g., depth/edge maps).
As a consequence, even the prevalent approaches struggle to generate high-fidality images semantically coherent to text prompts~\citep{li2024controlar, li2024controlvar}. As illustrated in Fig.~\ref{fig:figure1}, ControlAR~\cite{li2024controlar} demonstrates limited semantic comprehension of the textual prompt \textit{``A Jennifer Lopez film shot"},  relying primarily on the canny edge map to reconstruct regional details. Consequently, while the generated image maintains layout consistency with the control signals, its misaligned with the text prompt and generated artifacts.   

To bridge this substantial semantic gap, we leverage Multimodal Large Language Models (MLLMs) as a promising solution. Compared with controllable generation models, MLLMs \citep{shen2025vlm, guo2025deepseek, shu2024llava} excel at performing logical reasoning based on fundamental visual inputs~\citep{deng2025boosting, yang2025r1, zhang2025r1}. By introducing the MLLM as a sophisticated interpreter to uncover the latent semantics embedded in control images and enrich text prompts before generation, the controllable image generator can overcome the limitations of sparse textual semantic guidance, thereby enhancing generation quality. However, adapting an MLLM to the realm of controllable image generation is a non-trivial task. Unlike natural images, which are rich in color, texture, and contextual cues, control images (e.g., depth/edge maps) are structurally sparse, presenting significant challenges for semantic extraction \citep{huang2024controllable}. This discrepancy between control signals and natural images further complicates the faithful generation of the target image.

Building on the above analysis, in this paper, we propose \textbf{ControlThinker}, a novel framework that enhances controllable image generation through improved visual reasoning. As shown in Fig.~\ref{fig:figure1}, our ControlThinker employs a ``comprehend-then-generate" paradigm to effectively mine the latent semantics of control images. Specifically, rather than directly feeding the control image and text prompt into the image generator, we first leverage the modern MLLM, Qwen2.5-VL-7B~\cite{bai2025qwen25vltechnicalreport}, as an interpreter to comprehend the latent semantics in the control image and the input text prompt. However, in practice (Fig.~\ref{fig:figure3}), zero-shot prompting of the MLLM often fails to extract reliable latent semantics, owing to the inherently abstract, edge-only nature of control images (e.g., Canny edge maps).
To incentivize the visual reasoning capability in MLLM for comprehending control images, we design a data construction pipeline and propose \textbf{ControlThink-50K}, a high-quality visual reasoning CoT dataset specifically for control images. Meanwhile, we develop a two stage training paradigm to fine-tune the MLLM using both supervised (SFT) and reinforcement (RFT) techniques. After training, the MLLM can generate insightful chains of thought and produce enriched text prompts with dense semantics. Afterwards, such enriched text prompts, together with the control image, are fed into the generator to synthesize the target image. Despite not tuning the generator, overall performance improves significantly, demonstrating that our ControlThinker effectively unlocks the potential of pre-trained generation models. Additionally, while the enriched text prompts exhibit significantly increased semantic density, they are still influenced by the uncertainty stemming from the inherent lack of details in control images. To address this, we further expand the exploration of visual reasoning by sampling multiple reasoning trajectories from the MLLM and selecting the optimal one using a metric-based output reward model (ORM).

In summary, our contributions are three-folds: 
(1) We propose ControlThinker, a novel ``comprehend-then-generate" framework that introduces MLLM as a wise semantic interpreter to mine latent semantics from control images and enhance text prompts before image generation. (2) We propose ControlThink-50k, a high-quality visual reasoning CoT dataset and fine-tune the MLLM on with a two-stage supervised (SFT) and reinforcement learning (RFT) strategies. Additionally, we address visual reasoning uncertainty with a metric-based ORM to further boost performance. (3) Extensive experiments demonstrate that ControlThinker significantly improves performance across multiple control types, achieving superior visual quality and semantic consistency without modifying the underlying image generators, confirming the effectiveness of our approach.

%% file: sec/2_related_work.tex
\section{Related Work}

\subsection{Controllable Image Generation}
Traditional text-to-image approaches \citep{peebles2023scalable, ho2020denoising, dhariwal2021diffusion} relying solely on textual prompts often fall short in satisfying users' intentions of fine-grained generation. To enable precise control over pre-trained text-to-image diffusion models, additional visual control signals including Canny edges, HED boundaries, depth maps, and segmentation masks have been introduced as generation conditions. For instance, ControlNet \citep{zhang2023adding} integrates these visual conditions through an auxiliary parallel branch created by duplicating the pre-trained model. However, this extra module notably increases computational cost overhead. In contrast, T2I-Adapter \citep{mou2024t2i} offers a more efficient alternative with lightweight adapters, but sacrifices encoding capabilities for control images. More recently, ControlNet++ \citep{li2024controlnet++} addresses these limitations through a novel training paradigm, employing pixel-level cycle-consistency loss and efficient reward fine-tuning strategies to achieve more precise conditional control. Meanwhile, researchers have begun exploring multi-condition control \citep{zhao2023uni, hu2023cocktail, qin2023unicontrol} and instance-level control generation \citep{wang2024instancediffusion, zhou2024migc}. Recently, ControlAR \citep{li2024controlar} explores efficient controllable image generation for the autoregressive image generation model and designs a control information encoding module that operates directly on input tokens. Nevertheless, current methods predominantly concentrate on improving the model's capability to process textual and visual control signals, largely overlooking the intrinsic structural information inherent in control images. This structural information holds substantial potential for enriching the expressiveness of input textual prompts, thereby enhancing image generation quality from the very input stage.

\begin{figure*}[t] 
  \centering 
  \includegraphics[width=\textwidth]{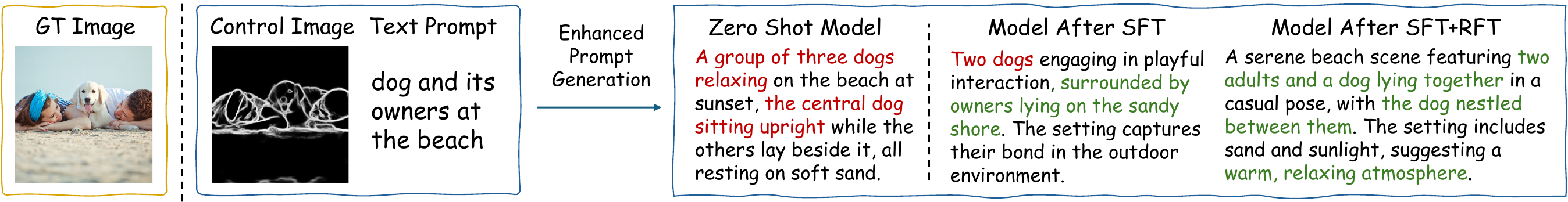} 
  \caption{Comparison of enhanced prompts from different models including zero shot model, model after SFT, and model after SFT and RFT. } 
  \label{fig:figure3} 
  \vspace{-6mm}
\end{figure*}


\begin{figure}[t] 
  \centering 
  \includegraphics[width=\linewidth]
  {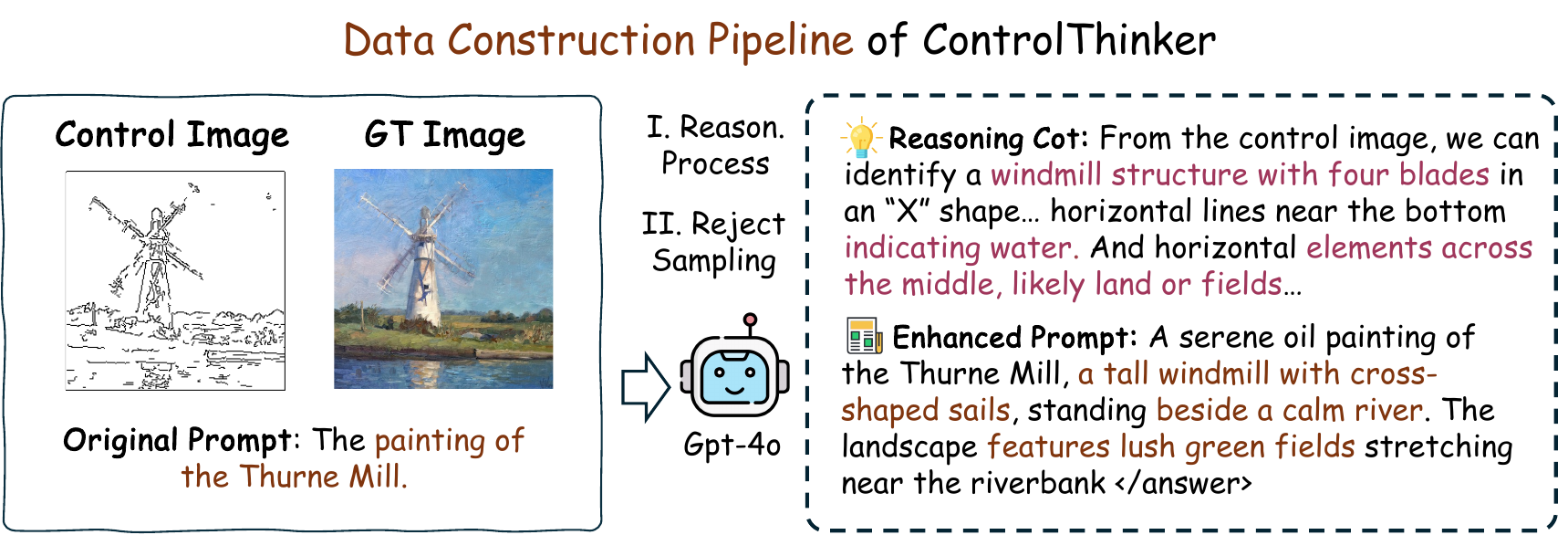} 
  \caption{The Data Construction Pipeline of ControlThinker.} 
  \label{fig:data_construct} 
  \vspace{-5mm}
\end{figure}

\subsection{Multimodal reasoning in image generation}
The emergence of multimodal large language models (MLLMs) \citep{bai2025qwen25vltechnicalreport, li2024llava} with advanced reasoning capabilities has inspired research on improving image generation quality through multimodal reasoning. For example, PARM++ \citep{guo2025can} employs a fine-tuned reward model at test-time to select generation paths aligned with text inputs, enhancing text-image coherence but unsuitable for autoregressive VQ-VAE models. GoT \citep{fang2025got} uses bounding boxes to plan object layouts, supporting precise image generation and editing. ImageGen-CoT \citep{liao2025imagegen} utilizes iterative prompt refinement for semantic alignment. However, in controllable generation tasks, control images inherently contain richer layout details than bounding boxes, making intermediate bounding boxes unnecessary. Effective multimodal reasoning thus requires analyzing these detailed layouts to enrich textual prompts with additional high-level semantics and low-level visual details, providing stronger control signals—an aspect ImageGen-CoT overlooks due to its semantic-centric approach.

%% file: sec/3_method.tex
\definecolor{ForestGreen}{rgb}{0, 0.69, 0.31}
\definecolor{NavyBlue}{rgb}{0, 0.44, 0.75}
\newcommand{\colorbold}[2][black]{\textcolor{#1}{\textbf{#2}}}
\newcommand{\hgreen}[1]{\textcolor{ForestGreen}{\textbf{#1}}}

\section{Method}
\subsection{Preliminary}
Firstly, we briefly review the basic formulation of the controllable image generation task. Formally, taking a text prompt $P_{orig}$ and a control image $I_{control}$ (e.g, depth/edge maps, segmentation masks) as input, the controllable image generation aims at generating a target image $I_{target}$ as formulated below:
\begin{equation}
    I_{\text{target}} = G\left(P_{\text{orig}}, I_{\text{control}}\right)
\end{equation}
where $G$ is the generation model. However, since $P_{\text{orig}}$ typically conveys only coarse or underspecified semantics, it offers limited guidance for region-level synthesis. Consequently, the generation process tends to over-rely on $I_{\text{control}}$ to infer fine-grained semantic details of $I_{\text{target}}$, often resulting in semantics distortions in localized areas as discussed in Sec.\ref{sec:intro}.

\subsection{ControlThinker}
Our framework consists of three primary components. Firstly, we design a data construction pipeline and curate ControlThink-50K, a high-quality visual reasoning dataset comprises both the reasoning CoT on control image and the original prompt, and the enhanced prompt with rich semantics (Fig.~\ref{fig:data_construct}). Besides, as illustrated in Fig.~\ref{fig:figure2}, we introduce a two-phase training strategy and inference time scaling techniques that work in sequence to enhance visual reasoning capabilities of the MLLM and the quality of generated images. Specifically, our training strategy comprises supervised fine-tuning (SFT) to establish fundamental reasoning capabilities, followed by reinforcement fine-tuning (RFT), during which the policy model generates multiple reasoning trajectories. These trajectories are assessed with verifiable rewards and optimized through Group Preference Optimization (GRPO). Moreover, our inference scaling techniques effectively generate multiple reasoning trajectories and images and selecting the optimal one based on image quality reward and layout consistency reward. Collectively, these methodologies constitute a novel "comprehend-then-generate" paradigm, effectively bridging the semantic gap between sparse textual prompts and target images, significantly enhancing image generation quality without necessitating alterations to generator.

\subsubsection{Data Construction Pipeline of ControlThinker}
\label{sec:data_curation}
To enable effective visual reasoning from control images, we design a data construction pipeline that encourages the MLLM to generate reasoning CoT and enhanced text prompts using powerful GPT-4o. Specifically, (1) we prompt GPT-4o~\cite{achiam2023gpt} to analyze and identify layout structures present in the control image. This spatial information, together with the original text prompt, is employed to enrich semantics of the text prompt, including objects and spatial relationships not explicitly mentioned in the original prompt. Meanwhile, to address the inherent limitations of control images (which lacks color and background information), we provide ground truth images as reasonable references about missing details. Moreover, (2) to further improve data quality, we apply sample-level rejection sampling. Firstly, we discard examples that does not conform to the required format ``\texttt{<think></think><answer></answer>}''. Second, we utilize GPT-4o to perform quality check and filter out reasoning CoT that perform analysis on the ground-truth (GT) image, since the GT image is unavailable at inference time.” This procedure ensure the quality of the data. Finally, we construct ControlThink-50K with this pipeline. We denote ControlThink-50K as $D\mathcal{=}\{(q^i, o^i)\}_{i=1}^N$ , where $q^i$ is the question for visual reasoning, and $o^i$ is the response containing the chain-of-thought reasoning process and the resulting enhanced prompt $P_{enh}^i$.


\subsubsection{Two-Phase Training}
To fully harness the visual reasoning capabilities of the MLLM for control images, we implement a two-phase training strategy. The first phase involves supervised fine-tuning (SFT) on ControlThink-50K to establish basic reasoning capabilities for control images. The second phase employs reinforcement fine-tuning (RFT) to evolve the model's reasoning abilities through exploring different reasoning trajectories, guided by well-designated verifiable rewards.

Before implementing our two-phase chain-of-thought training strategy, we conducted preliminary experiments to examine the limitations of the zero-shot prompting MLLM. With zero-shot inference, the model consistently misunderstood the content of control images owing to their inherently abstract, edge-only structure. As demonstrated in Fig.~\ref{fig:figure3}, the zero-shot model misinterpreted a scene with two adults and one dog as ``three dogs'', and incorrectly asserted that ``the central dog was sitting upright''. Therefore, we first adopt SFT to address this problem.

\begin{table*}[t]
\centering
\small
\setlength{\tabcolsep}{3pt}
\caption{\label{tab:t2i_consistency}\textbf{Layout consistency of T2I controllable generation.}  ``$\uparrow$'' or ``$\downarrow$'' indicate lower or higher values are better. ``-'' denotes that the method does not release a model for testing. The results are conducted on $512\times512$ resolution. ``*" denotes that the performances are evaluated using the official open-source checkpoint, which is employed as our baseline.}
\label{tab:sota_cosis}
\begin{tabular}{lccccc}
\toprule
\multirow{4}{*}{Method} & Seg & Canny & Hed  & Lineart & Depth  \\ \cmidrule{2-6} 
                        & mIoU $\uparrow$       & F1-Score $\uparrow$      & SSIM $\uparrow$ & SSIM $\uparrow$    & RMSE $\downarrow$  \\ \cmidrule{2-6} 
                        & COCOStuff       & MultiGen-20M      & MultiGen-20M & MultiGen-20M    & MultiGen-20M  \\ \midrule
GLIGEN~\cite{li2023gligen}           & -               & 26.94            & -       & -              & 38.83        \\
T2I-Adapter~\cite{mou2024t2i}        & -               & 23.65            & -           & -              & 48.40        \\
Uni-ControlNet~\cite{zhao2023uni}        & -               & 27.32            & 69.10       & -              & 40.65        \\
UniControl~\cite{qin2023unicontrol}            & -               & 30.82            & 79.69       & -              & 39.18        \\
ControlNet~\cite{zhang2023adding}            & 27.46           & 34.65            & 76.21       & 70.54          & 35.90        \\
ControlNet++~\cite{li2024controlnet++}          & 34.56           & 37.04            & 80.97       & \textbf{83.99}          & \underline{28.32}        \\

ControlAR* \tablefootnote{The open-source checkpoint is unified across multiple tasks, whereas the reported checkpoints in the paper were trained separately for each task.}~\cite{li2024controlar}            & \underline{37.90}           & \underline{37.30}            & \underline{83.52}       & 77.66          & 29.48        \\
\rowcolor{gray!20}ControlThinker (Ours)                & \textbf{40.07}           & \textbf{37.69}            & \textbf{84.24}       & \underline{78.22}         & \textbf{24.83}       \\ 
\bottomrule
\end{tabular}
\end{table*}

\begin{table*}[t]
\centering
\small
\setlength{\tabcolsep}{3pt}
\caption{\label{tab:t2i_fid}\textbf{FID of T2I controllable generation.} ``-'' denotes that the method does not release a model for testing. Our ControlThinker achieves significant FID improvements. ``*" shares the same meaning as in  Tab.~\ref{tab:sota_cosis}.}
\label{tab:sota_fid}
\begin{tabular}{lccccc}
\toprule
\multirow{2.5}{*}{Method} & Seg & Canny & Hed  & Lineart & Depth  \\ \cmidrule{2-6}  & COCOStuff       & MultiGen-20M      & MultiGen-20M & MultiGen-20M    & MultiGen-20M  
                          \\ \midrule
GLIGEN~\cite{li2023gligen}                   & -               & 18.89            & -       & -              & 18.36        \\
T2I-Adapter~\cite{mou2024t2i}             & -               & \underline{15.96}            & -           & -              & 22.52        \\
Uni-ControlNet~\cite{zhao2023uni}          & -               & 17.14            & 17.08       & -              & 20.27        \\
UniControl~\cite{qin2023unicontrol}              & -               & 19.94            & 15.99       & -              & 18.66        \\
ControlNet~\cite{zhang2023adding}             & 21.33           & \textbf{14.73}            & 15.41       & 17.44          & 17.76        \\
ControlNet++~\cite{li2024controlnet++}           & 19.29           & 18.23            & 15.01       & \underline{13.88}         & \underline{16.66}        \\
ControlAR*~\cite{li2024controlar}                  & \underline{14.31}           & 24.08          & \underline{12.52}       & 14.39         & 17.38        \\ 
\rowcolor{gray!20}ControlThinker (Ours)                  & \textbf{14.09}           & 19.76            & \textbf{11.39}       & \textbf{12.68}         &  \textbf{16.07}       \\ 
\bottomrule
\end{tabular}
\vspace{-2mm}
\end{table*}

\textbf{SFT.}\quad Our first training phase employs supervised fine-tuning on ControlThink-50K to establish basic visual perception and reasoning capabilities.
Let $D=\{(q_i,o_i)\}_{i=1}^{N}$ with $o_i=(o_{i,1},\ldots,o_{i,T_i})$.
The SFT loss is
\begin{equation}
\label{eq:sft-sum-unmasked}
\mathcal{L}_{\text{SFT}}
= - \frac{1}{N}\sum_{i=1}^{N}\frac{1}{T_i}
\sum_{t=1}^{T_i}\log \pi_{\theta}\!\left(o_{i,t}\mid q_i,\,o_{i,<t}\right),
\end{equation}
where $\pi_{\theta}$ is the policy model parameterized by $\theta$.

As demonstrated in Fig.~\ref{fig:figure3}, while the model after SFT could recognize ``the owners'', it partially misread the control image as containing ``two small dogs''. This necessitates our second phase training strategy.

\textbf{RFT.}\quad The second phase reinforcement fine-tuning is where the model's reasoning abilities truly evolve. During this phase, for each question $q$, we generate multiple reasoning trajectories with diverse interpretations of the visual content. From each trajectory, we extract the final enhanced prompt $\{P_{enh}^1, P_{enh}^2, ..., P_{enh}^N\}$ and evaluate its alignment with the ground truth image $I_{gt}$ using CLIP scores as a verifiable reward metric. The reward for each enhanced prompt $P_{enh}^i$ is calculated as:
\begin{equation}
    R_{align} = \text{CLIP}(I_{gt}, P_{enh}^i),
\end{equation}
where $I_{gt}$ is the ground-truth image and CLIP computes the image-text similarity score. 

Following DeepSeek-R1 \citep{guo2025deepseek}, we introduce the format reward which verifies whether each response conform to the required format  ``\verb|<think></think><answer></answer>|''. Responses adhering to this format receive a reward of 1; otherwise, they receive 0. The final reward combines both the CLIP similarity score and this format adherence reward:

\begin{figure*}[t] 
  \centering 
  \includegraphics[width=0.8\linewidth]{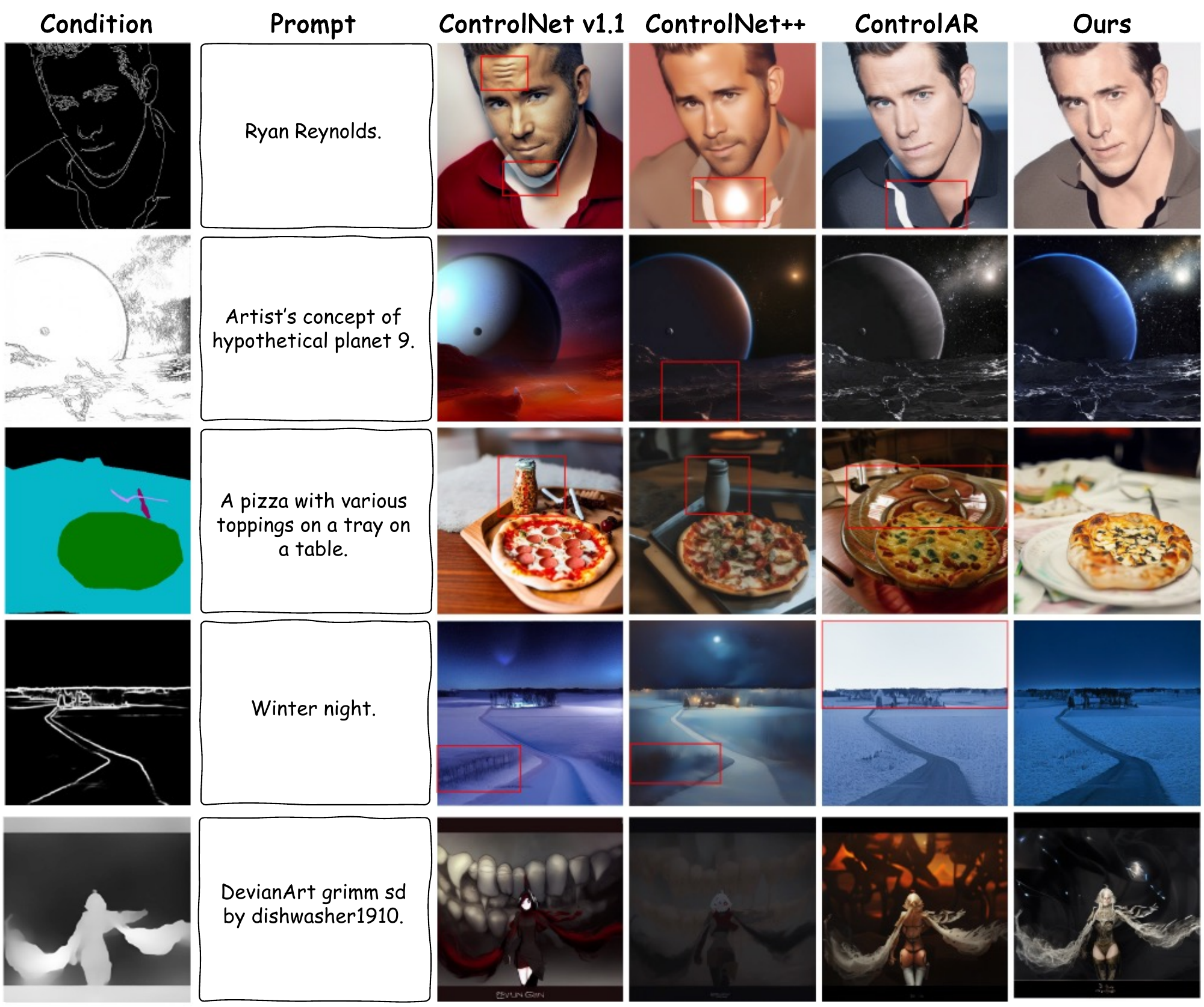} 
  \caption{Qualitative comparison with state-of-the-art approaches across different condition types.} 
  \label{fig:qual} 
  \vspace{-3mm}
\end{figure*}

\begin{equation}
    R(q,o^i) = R_{align} + R_{format}
\end{equation}
This approach allows the model to learn from its own reasoning variations, identifying which reasoning pathways produce the most accurate and coherent dense semantic information. To systematically compare reasoning pathways, we implement Group Preference Optimization (GRPO), the relative advantage of each response within the group is computed as:
\begin{equation}
    A_i = \frac{R_i - \text{mean}(\{R_1, \dots, R_G\})}{\text{std}(\{R_1, \dots, R_G\})},    
\end{equation}
where $R_j = R(q, o^j)$ for $j = 1,2,...,G$.
To maintain training stability and prevent excessive divergence from the reference model $\pi_{ref}$, we incorporate a KL divergence penalty weighted by coefficient $\beta$ into our objective function. Finally, the objective function is calculated as:
\begin{equation}
\resizebox{\linewidth}{!}{$
\mathcal{L}_{\text{RFT}} = \mathbb{E}_{q \sim D} [ \sum_{i=1}^G A_i \log \pi_{\theta}(o^i|q) 
- \beta \text{KL}[\pi_{\theta}(o^i|q) \parallel \pi_{\text{ref}}(o^i|q)] ]$
}
\end{equation}
As evidenced in Fig.~\ref{fig:figure3}, our fully trained model demonstrates sophisticated understanding of implicit visual details from the control image (``two
adults and a dog lying together'' and ``dog nestled
between them'').


\subsubsection{Inference Time Scaling}
Due to the inherent limitations of control images, such as missing color information, background context, and other fine-grained details, the enhanced prompts, while semantically accurate, may still carry a degree of uncertainty. To mitigate this, we introduce an inference-time scaling strategy designed to fully leverage the model’s visual reasoning capabilities.
Specifically, we sample multiple enhanced prompts for the same control image and original text. Each enhanced prompt is used to generate a candidate image, which is then evaluated using our dual-metric assessment framework that accounts for both high-level quality and semantic alignment and low-level layout consistency to the control image.

For high-level quality and semantics, we adopt PickScore~\citep{kirstain2023pick} to quantitatively assess the image quality as well as the semantic alignment between each candidate image and the original prompt. Simultaneously, for low-level layout consistency, we calculate the LPIPS score~\citep{zhang2018unreasonable} between the control map extracted from each candidate image and the original control image, quantifying structural consistency. Formally, for a set of $K$ candidate images $\{I_1, I_2, ..., I_K\}$ generated from enhanced prompts $\{P_{enh}^1, P_{enh}^2, ..., P_{enh}^K\}$ and the control image, we compute a combined reward score:
\begin{equation}
    R(I_k) = \text{PS}(I_k, P_{orig}) - \text{LPIPS}(C(I_k), I_{control})
\end{equation}
where $PS$ indicates PickScore, $P_{orig}$ is the original prompt, $C(I_k)$ is the control map extracted from the generated image $I_k$, $I_{control}$ is the original control image. We select the candidate image with the highest reward:
\begin{equation}
    I^* = \arg\max_{I_k} R(I_k)
\end{equation}
This selected image excels in both fidelity, text alignment and structural adherence to the control signal.

\begin{table*}[t]
\centering
\caption{\textbf{Comparison between our method and other text prompt enhancement approaches.} ``\textcolor{red}{$\uparrow$}'' and ``\textcolor{red}{$\downarrow$}'' indicate lower and higher values are better, respectively. Values in \textcolor{red}{red} indicate improvements, while \hgreen{green} indicates degradations.}
\label{tab:pmt_enhance}
\scalebox{0.8}{
\begin{tabular}{ccccccccccc}
\toprule
\multirow{3}{*}{Method} & \multicolumn{2}{c}{Seg}       & \multicolumn{2}{c}{Canny}        & \multicolumn{2}{c}{Hed}          & \multicolumn{2}{c}{Lineart}      & \multicolumn{2}{c}{Depth}        \\ \cmidrule{2-11} 
                        & \multicolumn{2}{c}{COCOStuff} & \multicolumn{2}{c}{MultiGen-20M} & \multicolumn{2}{c}{MultiGen-20M} & \multicolumn{2}{c}{MultiGen-20M} & \multicolumn{2}{c}{MultiGen-20M} \\ \cmidrule{2-11} 
                        & FID \textcolor{red}{$\downarrow$}           & mIoU \textcolor{red}{$\uparrow$}           & FID \textcolor{red}{$\downarrow$}            & F1-Score \textcolor{red}{$\uparrow$}        & FID \textcolor{red}{$\downarrow$}             & SSIM \textcolor{red}{$\uparrow$}           & FID \textcolor{red}{$\downarrow$}             & SSIM \textcolor{red}{$\uparrow$}           & FID \textcolor{red}{$\downarrow$}             & RMSE \textcolor{red}{$\downarrow$}           \\ \midrule
ControlAR               & 14.31         & 37.90         & 24.08          & 37.30           & 12.52           & 83.52          & 14.39           & 77.66          & 17.38           & 29.48          \\ \hline
+ T2I-R1~\cite{jiang2025t2i}                & 15.74         & 35.62         & 21.50          & 37.29           & 11.84           & 80.10          & 12.89           & 75.87          & 17.86           & 29.37          \\
\rowcolor{gray!20}$\Delta$                   & \colorbold[ForestGreen]{+1.43}         & \colorbold[ForestGreen]{-2.28}         & \colorbold[red]{-2.58}          & \colorbold[ForestGreen]{-0.01}           & \colorbold[red]{-0.68}           & \colorbold[ForestGreen]{-3.42}          & \colorbold[red]{-1.50}           & \colorbold[ForestGreen]{-1.79}          & \colorbold[ForestGreen]{+0.48}           & \colorbold[red]{-0.11}          \\
+ GPT-4o~\cite{achiam2023gpt}                & 17.03         & 35.24         & 21.08          & 37.40           & 12.04           & 83.72          & 13.17           & 76.02          & 19.62           & 29.42          \\
\rowcolor{gray!20}$\Delta$                   & \colorbold[ForestGreen]{+2.72}         & \colorbold[ForestGreen]{-2.66}         & \colorbold[red]{-3.00}          & \colorbold[red]{+0.10}           & \colorbold[red]{-0.48}           & \colorbold[red]{+0.20}          & \colorbold[red]{-1.22}           & \colorbold[ForestGreen]{-1.64}          & \colorbold[ForestGreen]{+2.24}           & \colorbold[red]{-0.06}          \\
+ Ours                  & 14.09         & 40.07         & 19.76          & 37.69           & 11.39           & 84.24          & 12.68           & 78.22          & 16.07           & 24.83          \\
\rowcolor{gray!20}$\Delta$                   & \colorbold[red]{-0.22}         & \colorbold[red]{+2.17}         & \colorbold[red]{-4.32}          & \colorbold[red]{+0.33}           & \colorbold[red]{-1.13}           & \colorbold[red]{+0.72}          & \colorbold[red]{-1.71}           & \colorbold[red]{+0.56}          & \colorbold[red]{-1.31}           & \colorbold[red]{-4.54}          \\ \bottomrule
\end{tabular}}
\end{table*}

%% file: sec/4_experiment.tex
\section{Experiments}

\subsection{Experimental Setup}
\textbf{Datasets and Models.}\quad
We implement our ControlThinker framework using Qwen2.5-VL-7B as the base MLLM for visual reasoning. Our controllable image generation experiments are conducted using ControlAR as the generator backbone, which is an autoregressive text-to-image model built on LlamaGen. We evaluate our approach across five control signal types: segmentation masks, canny edges, hed edges, lineart edges, and depth maps. For experiments on the segmentation mask, we utilize the COCOStuff dataset. For the four remaining control types, we employ the MultiGen-20M dataset, which contains various control signals paired with high-quality images and corresponding text prompts.

\textbf{Implementation Details.}\quad For SFT, we adopt LoRA (rank=32, $\alpha$=64) with Adam optimizer (lr=5e-6, batch size=6) for one epoch. For RFT, we employed LoRA (rank=64, $\alpha$=128) with Adam (lr=1e-5, batch size=1), sampling 12 responses per question over 2,400 training steps. During inference scaling, we generated 10 enhanced prompts per data point, with each prompt generating one image and select the optimal one. All experiments are conducted on 8 NVIDIA A100 GPUs.

\textbf{Evaluation Metrics.}\quad
To assess controllable image generation capability, we adopt two metrics: layout consistency and FID. Layout consistency evaluates the structural fidelity between generated images and original control signals using task-specific metrics: mean Intersection-over-Union (mIoU) for segmentation masks, F1-Score for canny edges, Structural Similarity Index (SSIM) for hed and lineart edges, and Root Mean Square Error (RMSE) for depth maps. To evaluate overall generation quality, we utilize Fréchet Inception Distance (FID), which measures distributional similarity between generated and real images, with lower values indicating better image generation quality. Additionally, we employ PickScore to assess text-image alignment quality between generated prompts and real images.

\subsection{Comparison with State-of-the-art Methods}
Following the evaluation protocol of prior works~\cite{li2024controlar,li2024controlnet++}, we comprehensively compare our proposed method against ControlAR and other state-of-the-art baselines across two key dimensions: (1) layout consistency to the control images, and (2) distributional realism with respect to ground-truth images. The results for each are presented in Tab.~\ref{tab:sota_cosis} and Tab.~\ref{tab:t2i_consistency}, respectively.

As shown in Tab.~\ref{tab:t2i_fid}, our ControlThinker achieves state-of-the-art structural consistency across a range of conditional image generation tasks. Notably, it surpasses the well-trained ControlNet, its reinforcement learning enhanced counterpart ControlNet++, and the autoregressive baseline ControlAR in preserving layout consistency across Canny, Hed, Seg, and Depth control conditions. Quantitatively, ControlThinker demonstrates significant gains, evidenced by a substantial increase of 2.17 in mIoU on the COCOStuff dataset for segmentation task and a notable reduction of 3.49 in RMSE on the MultiGen-20M dataset for depth task. These results underscore the superior capability of our proposed ControlThinker in generating images that exhibit both strong adherence to the input conditions and enhanced structural integrity compared to existing state-of-the-art methods.

Meanwhile, as demonstrated in Tab.~\ref{tab:sota_fid}, our ControlThinker also achieves state-of-the-art Fréchet Inception Distance (FID) scores across Seg, Hed, Lineart, and Depth tasks, indicating the superior image quality of our generated outputs. Our consistent FID gains across different control settings, achieved without generator fine-tuning, show that our method supplies richer semantic information, helping the generator produce higher-quality images without compromising structural integrity as shown in Tab.~\ref{tab:sota_cosis}. The concurrent enhancement of both structural consistency and image quality underscores the effectiveness of ControlThinker in leveraging semantic guidance for improved image generation.

\subsection{Prompt-Enhancement Comparison}
To validate the effectiveness of our approach in mining latent semantics, we conduct a comprehensive comparison with state-of-the-art prompt-enhancement methods. As reported in Tab.~\ref{tab:pmt_enhance}, T2I-R1~\cite{jiang2025t2i} and GPT-4o~\cite{achiam2023gpt} consistently degrade layout consistency—measured by mIoU, F1-Score, SSIM, and RMSE—on both the segmentation and line-art tasks. 
\begin{table}[h]
\centering
\small
\caption{Ablation study of each design in our method and their combination.}
\label{tab:perform_breakdown}
\resizebox{0.85\columnwidth}{!}{%
\begin{tabular}{lcccc}
\toprule
\multirow{2.5}{*}{Method} & \multicolumn{2}{c}{Hed}  & \multicolumn{2}{c}{Lineart}  \\ \cmidrule{2-5} 
                        & SSIM$\uparrow$ & FID$\downarrow$ & SSIM$\uparrow$ & FID$\downarrow$  \\ \midrule
Baseline & 83.52 & 12.52 & 77.66 & 14.39 \\
+Zero Shot & 83.13 & 13.64 & 77.23 & 16.66 \\
+SFT & 83.65 & 12.11 & 77.67 & 13.39 \\
+SFT+RFT & 83.72 & \underline{11.78} & 77.72 & \underline{12.99} \\ 
+Inference Scaling & \underline{84.02} & 12.17 & \textbf{78.24} & 13.86 \\
ControlThinker & \textbf{84.24} & \textbf{11.39} & \underline{78.22} & \textbf{12.68} \\ \bottomrule
\end{tabular}
}
\vspace{-2mm}
\end{table}
Meanwhile, T2I-R1 diminishes layout alignment for Hed and line-art inputs, and GPT-4o achieves only limited gains. By contrast, our method improves all layout metrics, with notable benefits for segmentation and depth tasks. In terms of image fidelity, assessed via FID, T2I-R1 and GPT-4o yield improvements only for canny inputs but negatively affect segmentation and depth, indicating semantic drift from the intended control signals. Our method consistently improves FID across all conditions without modifying the generator, producing fewer artifacts and textures more closely aligned with the real-image distribution while preserving structural accuracy.

\subsection{Comprehensive Analysis}

\begin{table}[t]
\centering
\small
\caption{Influences of different SFT approaches over RFT.}
\label{tab:sft_rft}
\resizebox{0.85\columnwidth}{!}{%
\begin{tabular}{lcccc}
    \toprule
    \multirow{2.5}{*}{Method} & \multicolumn{2}{c}{Hed}  & \multicolumn{2}{c}{Lineart}  \\ \cmidrule{2-5} 
                            & SSIM$\uparrow$ & FID$\downarrow$ & SSIM$\uparrow$ & FID$\downarrow$  \\ \midrule
    Baseline  & 83.52 & 12.52 & 77.66 & 14.39 \\
    +RFT & 83.58 & 13.12 & 77.53 & 16.08 \\
    +SFT(w/o CoT)+RFT & \underline{83.67} & \underline{12.00} & \underline{77.67} & \underline{14.17} \\ 
    +SFT(w/ CoT)+RFT & \textbf{83.72} & \textbf{11.78} & \textbf{77.72} & \textbf{12.99} \\ \bottomrule
\end{tabular}
}
\vspace{-2mm}
\end{table}

\begin{figure}[t]
    \centering
    \includegraphics[width=\linewidth]{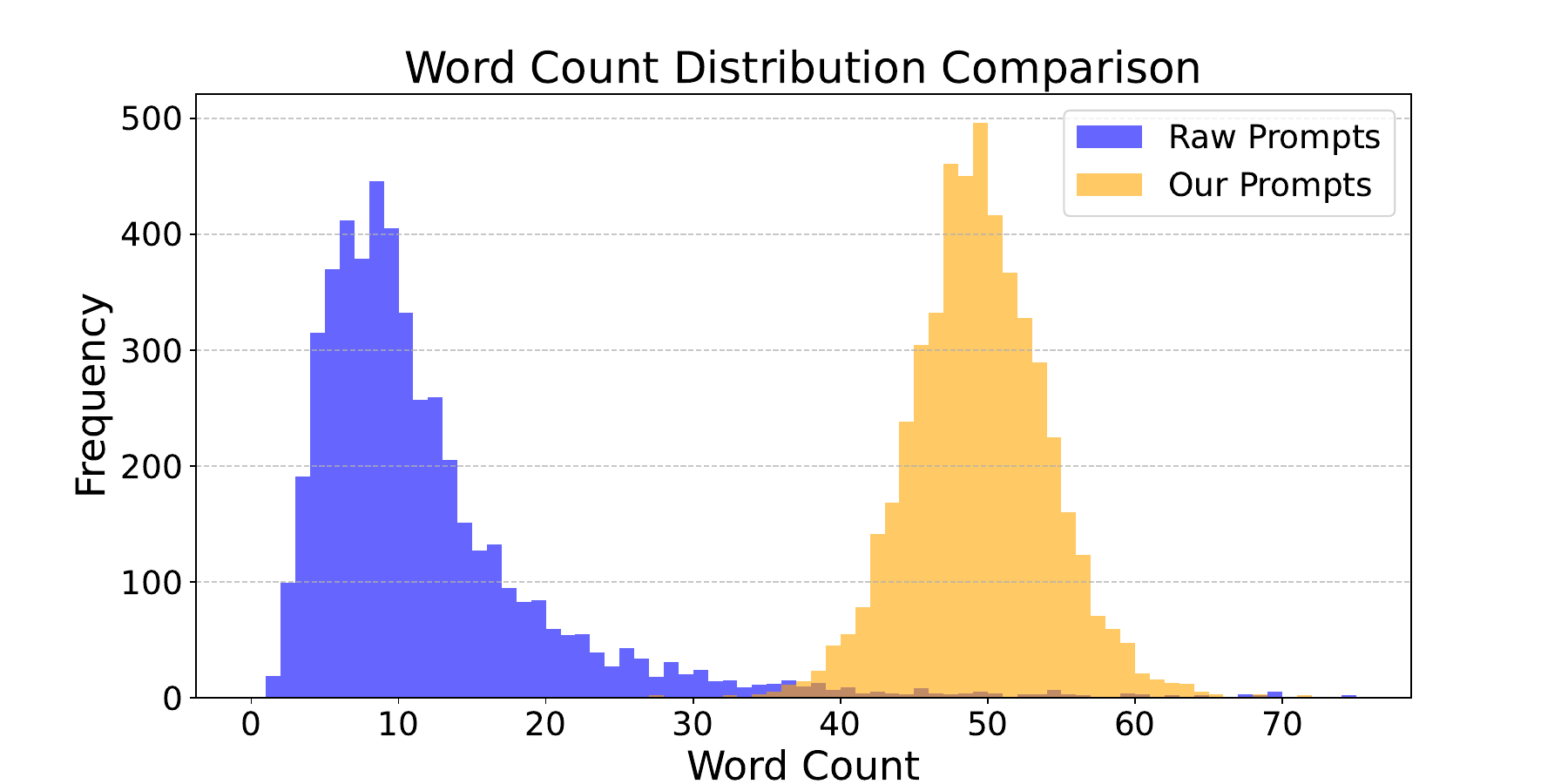}
    \caption{Length distribution of text prompts in the official dataset versus our curated dataset enriched with visual semantics.}
    \label{fig:word-count}
\vspace{-2mm}
\end{figure}

\noindent \textbf{Performances Breakdown.}\quad To systematically evaluate the effectiveness of each design in our proposed ControlThinker, we ablate their individual effects as shown in Tab.~\ref{tab:perform_breakdown}. We observe that both SFT and the combination of SFT+RFT consistently improve SSIM and notably reduce FID across Hed and Lineart styles. Besides, the inference scaling further boosts SSIM. Combining all techniques, our ControlThinker achieves the best overall performance.

\noindent \textbf{Comparison between Enhanced and Original Prompts.}\quad To characterize our enhanced prompts following data curation, we analyze their length distribution compared to the original text prompts from the official dataset (Fig.~\ref{fig:word-count}). Our analysis indicates that the enhanced prompts exhibit significantly increased length\footnote{Lengths are calculated only for final answers without CoT}, suggesting that they encapsulate more comprehensive semantic information.

\noindent \textbf{Effects of SFT vs. RFT.}\quad
Prior work suggests that applying domain-specific SFT before RFT can over-constrain exploration~\cite{liu2025visual,meng2025mm}. We examine this interplay in controllable generation (Tab.~\ref{tab:sft_rft}). RFT alone underperforms the baseline with higher FID and lower SSIM, indicating that the pretrained MLLM lacks sufficient control-image understanding, pushing RL toward biased, suboptimal trajectories. In contrast, SFT with CoT-style reasoning yields consistent, sizable gains, showing that structured reasoning supervision is key to correctly parsing control signals and producing semantically coherent outputs. The gap between CoT-training and plain training further underscores the value of explicit reasoning cues.

\noindent \textbf{Computational Cost of Inference Time Scaling.}\quad
As demonstrated in Tab.~\ref{tab:parallel-inference} we denote the original single-image generation time as the $1.0\times$ baseline, our parallel inference time scaling only increases end-to-end runtime to $1.48\times$ while keeping memory usage under 24GB. Overall, the additional $48\%$ time yields consistent improvements on key metrics, making the approach a worthwhile trade-off for controllable T2I generation.

\begin{table}[h]
\centering
\small
\caption{Computational cost analysis of our inference time scaling method under parallel inference. Stage I: Enhanced prompts generation. Stage II: Images generation. Stage III: Score computation. \textbf{Memory (GB)}: Peak GPU memory usage.}
\label{tab:parallel-inference}
\begin{tabular}{lcc}
\toprule
\textbf{Stage} & \textbf{Time Ratio} & \textbf{Memory (GB)} \\
\midrule
Original & 1.00$\times$ &  6.52 \\
\midrule
Stage I        & 0.25$\times$ & 19.36 \\
Stage II     & 1.18$\times$ & 21.60 \\
Stage III         & 0.05$\times$ &  4.01 \\
\midrule
\textbf{Total}        & 1.48$\times$ & 21.60 \\
\bottomrule
\end{tabular}
\vspace{-2mm}
\end{table}

\subsection{Qualitative Results Demonstration}
As demonstrated in Fig.~\ref{fig:qual}, our method generates results with higher visual fidelity and semantic accuracy. In the first row, other methods produce artifacts or distortions, while ours maintains portrait realism. In the third row, ControlNet and ControlNet++ wrongly add condiment bottles, and ControlAR generates an extra tray. Only our model accurately follows both the layout and the prompt, demonstrating better semantic understanding and structure control.



%% file: sec/5_conclusion.tex
\section{Conclusion}
In this paper, we present ControlThinker, a novel ``comprehend-then-generate'' framework that leverages a fine-tuned Multimodal Large Language Model (MLLM) to bridge the semantic gap in controllable image generation. By treating the MLLM as a visual reasoning engine, ControlThinker effectively extracts latent semantics from control images, enriching text prompts with denser guidance. Without modifying the image generator, our method achieves notable improvements in both layout consistency and semantic alignment. Extensive experiments across multiple control types confirm the effectiveness of our approach.

%% file: sec/X_suppl.tex
\clearpage
\appendix
\setcounter{page}{1}
\maketitlesupplementary

\section{Baseline Details}

In our comparative experiments, GLIGEN utilizes SD1.4 as the generative model, while T2I-Adapter, Uni-ControlNet, UniControl, ControlNet, and ControlNet++ adopt SD1.5 as the generative model.

For the prompt enhancement method we compare in our experiment section, T2I-R1 is adopted as the single-modal text enhancement baseline while GPT-4o is adopted as the multimodal understanding and prompt enhancing method.


\section{More Qualitative Comparison Results}
\begin{figure*}[t] 
  \centering 
  \includegraphics[width=0.9\linewidth]{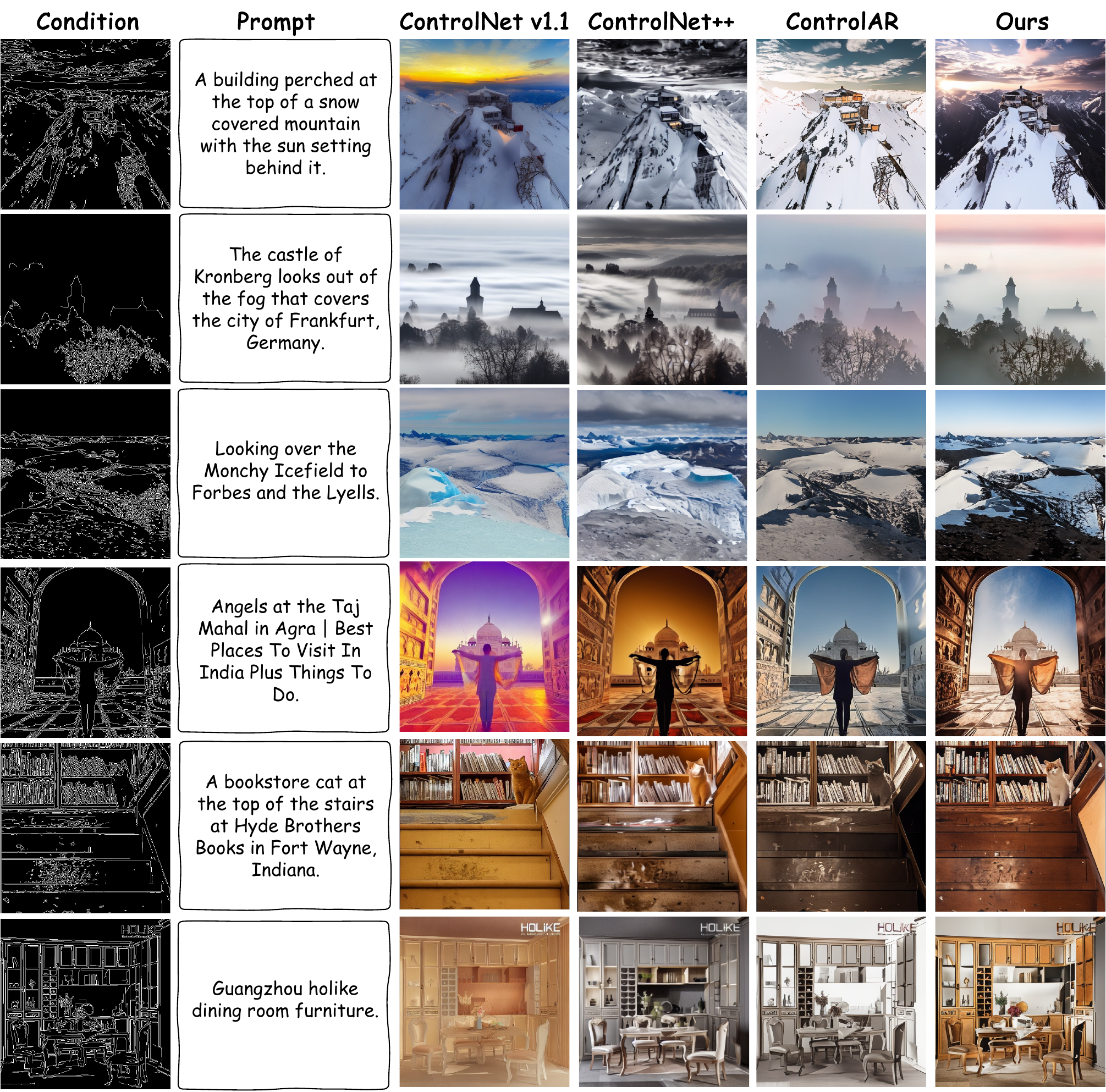} 
  \caption{Qualitative comparison with state-of-the-art approaches on Canny condition types.} 
  \label{fig:qual_canny} 
  \vspace{-3mm}
\end{figure*}

\begin{figure*}[t] 
  \centering 
  \includegraphics[width=0.9\linewidth]{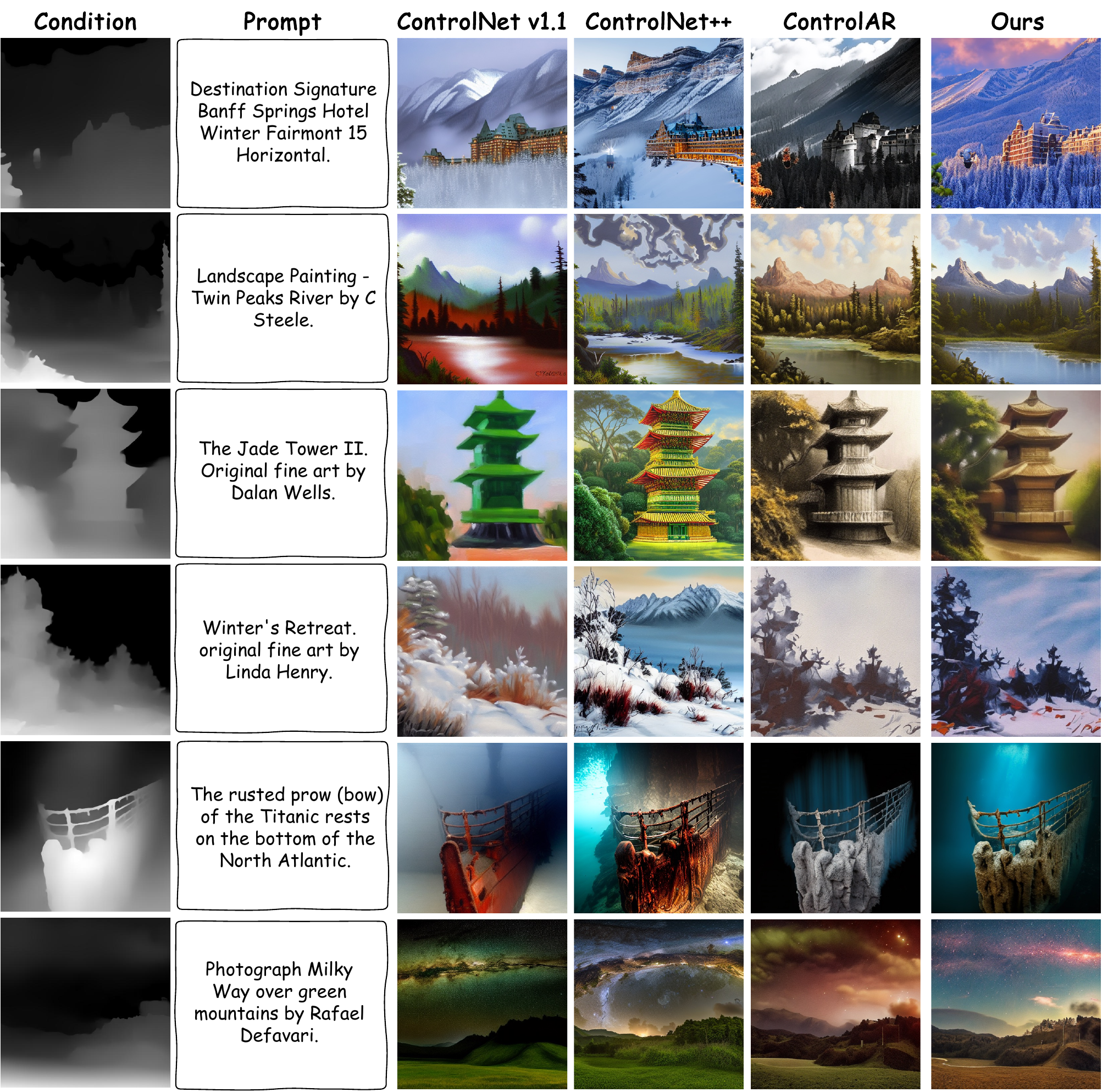} 
  \caption{More qualitative comparison with state-of-the-art approaches on Depth condition types.} 
  \label{fig:qual_depth} 
  \vspace{-3mm}
\end{figure*}

\begin{figure*}[t] 
  \centering 
  \includegraphics[width=0.9\linewidth]{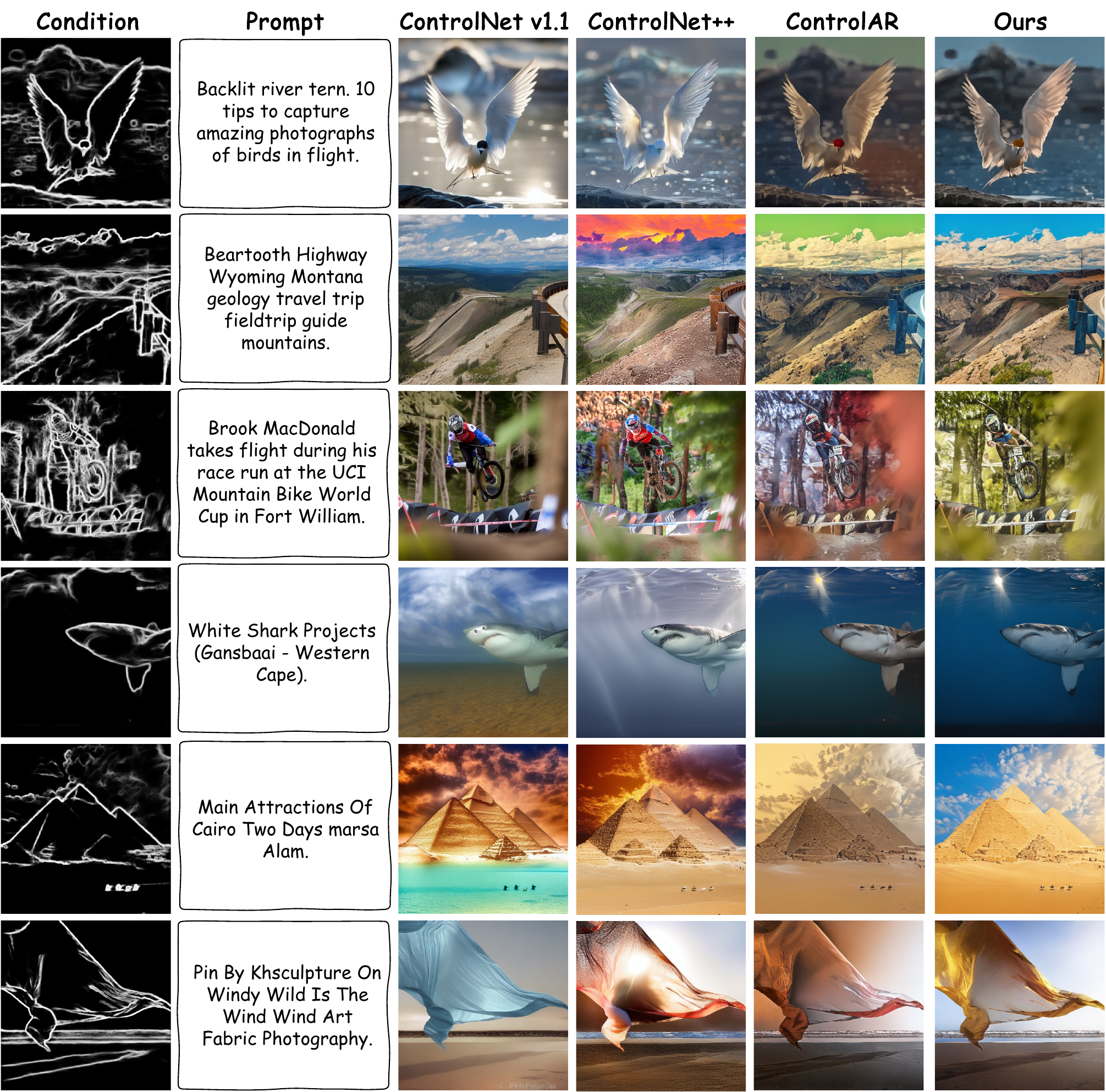} 
  \caption{More qualitative comparison with state-of-the-art approaches on Hed condition types.} 
  \label{fig:qual_hed} 
  \vspace{-3mm}
\end{figure*}

\begin{figure*}[t] 
  \centering 
  \includegraphics[width=0.9\linewidth]{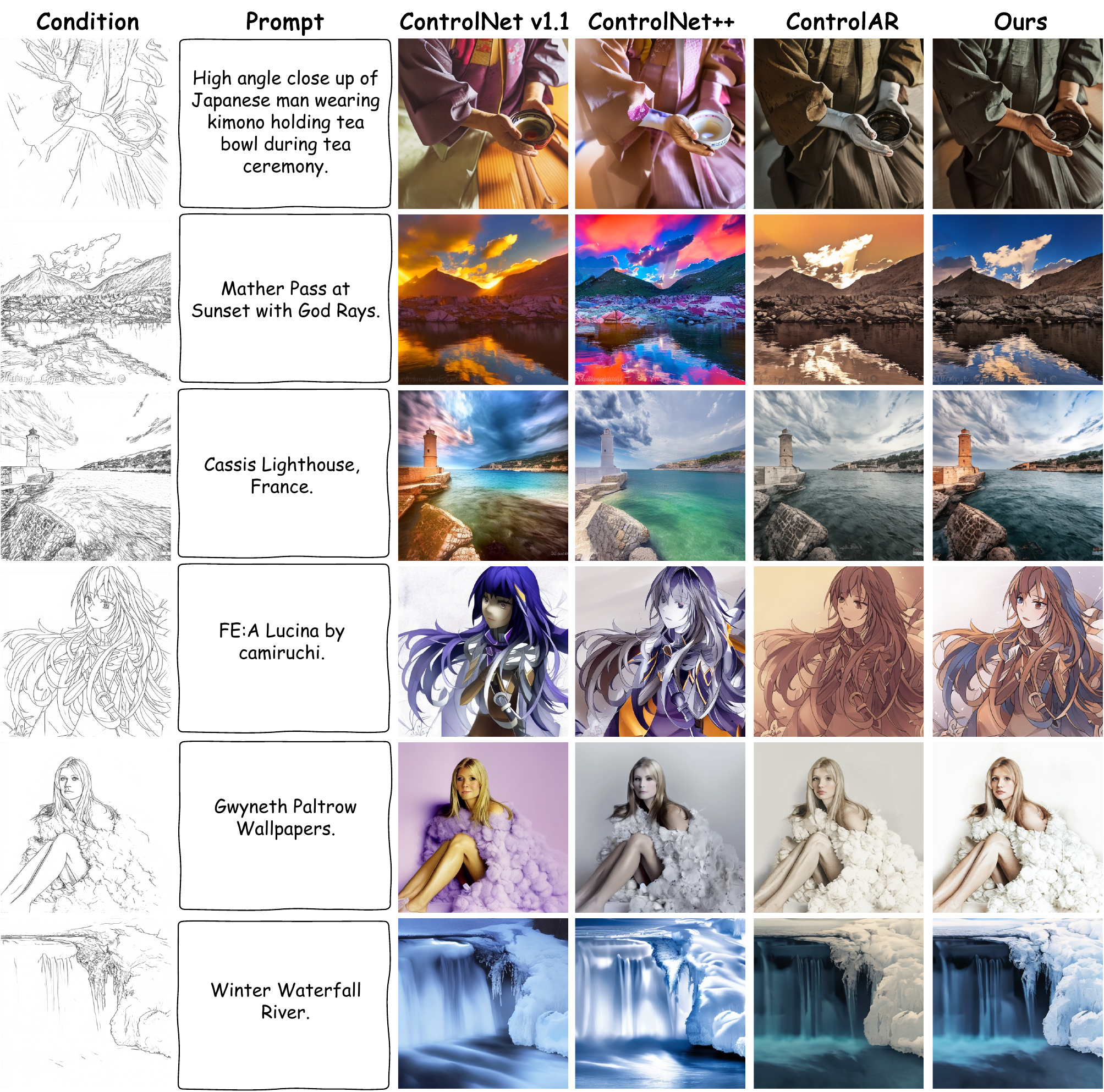} 
  \caption{More qualitative comparison with state-of-the-art approaches on Lineart condition types.} 
  \label{fig:qual_lineart} 
  \vspace{-3mm}
\end{figure*}

\begin{figure*}[t] 
  \centering 
  \includegraphics[width=0.9\linewidth]{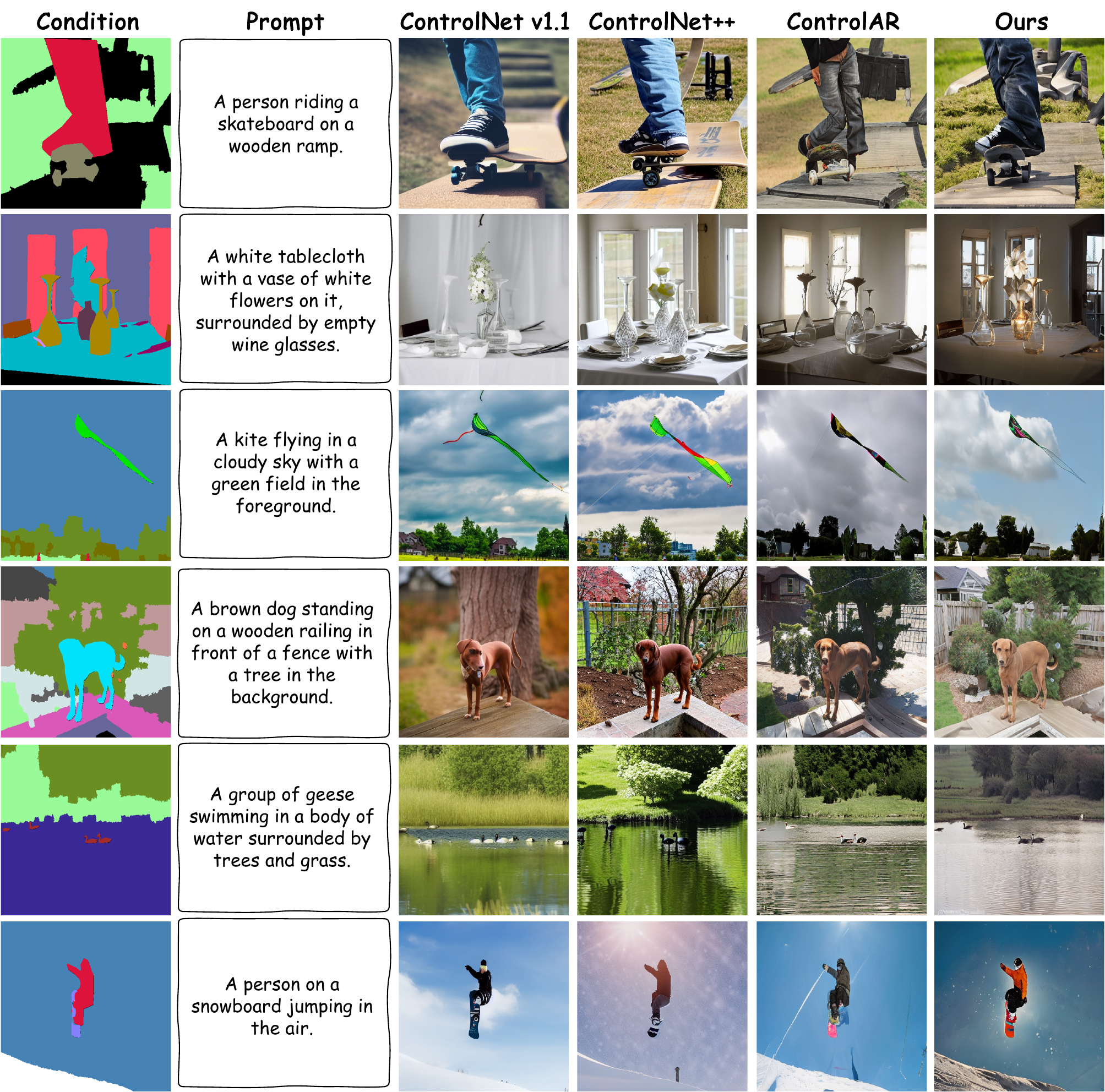} 
  \caption{More qualitative comparison with state-of-the-art approaches on Segmentation condition types.} 
  \label{fig:qual_seg} 
  \vspace{-3mm}
\end{figure*}

We provide additional qualitative comparisons for Canny task on Fig.~\ref{fig:qual_canny}, Depth task on Fig.~\ref{fig:qual_depth}, Hed task on Fig.~\ref{fig:qual_hed}, Lineart task on Fig.~\ref{fig:qual_lineart}, Segmentation task on Fig.~\ref{fig:qual_seg}.

\section{More Visualization of Enhanced Prompts Generated by ControlThinker.}
\begin{figure*}[t] 
  \centering 
  \includegraphics[width=0.9\linewidth]{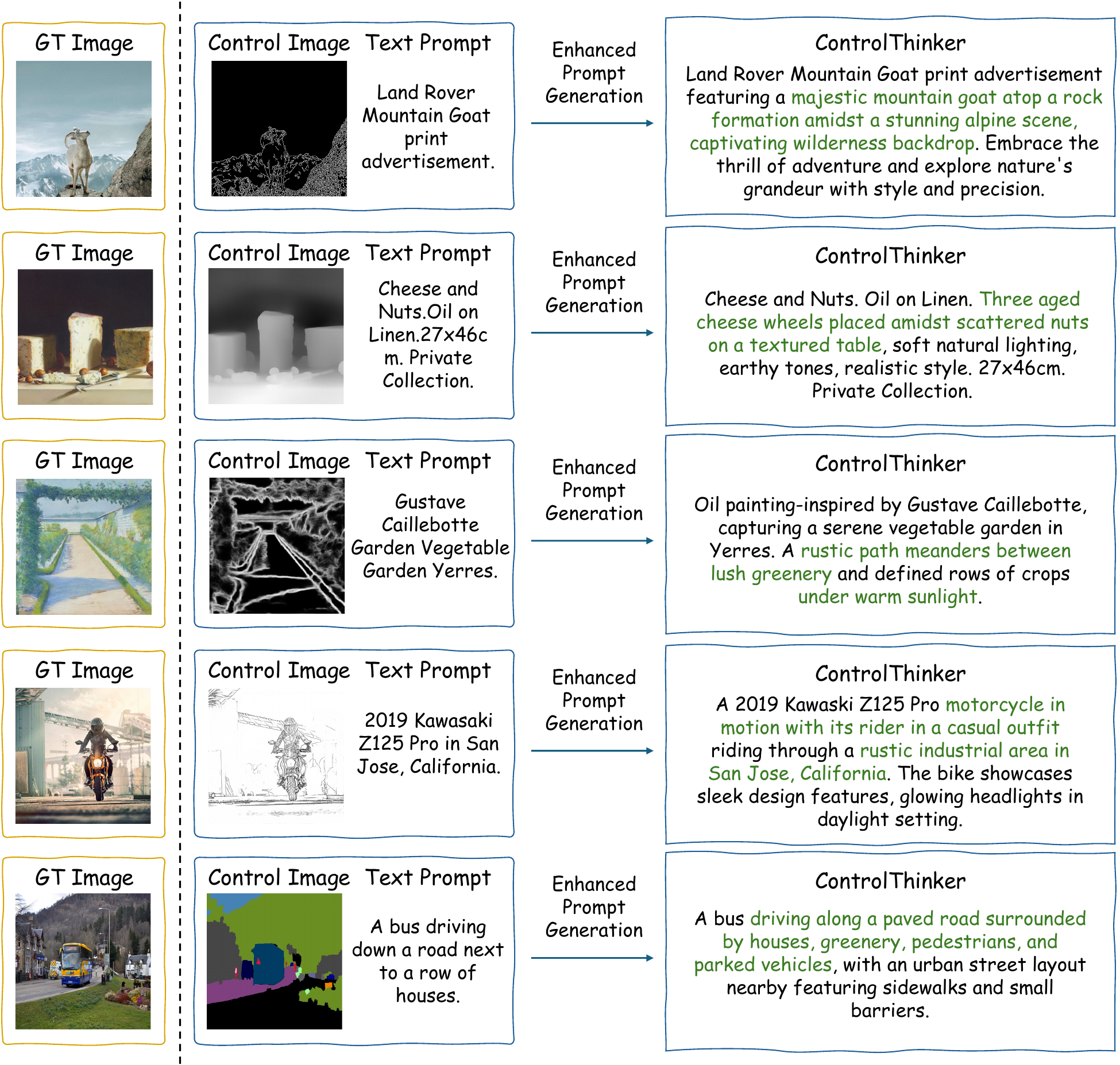} 
  \caption{Visualization of enhanced prompts generated by ControlThinker.} 
  \label{fig:reasoning_prompts_appendix} 
\end{figure*}

Fig.~\ref{fig:reasoning_prompts_appendix} demonstrates the enhanced prompts from ControlThinker, which accurately describe the images to generate.

\section{Prompts for the Data Construction and The Two Phase Training Process}
Tab.~\ref{tab:dataprompt} is the user prompt for the Data Construction. Tab.~\ref{tab:dataprompt2} and Tab.~\ref{tab:dataprompt3} are system prompt and user prompt for the Two Phase Training.

\clearpage

\begin{table*}[t]
    \centering
    \caption{Template of the user prompt for Data Construction. Condition types include hed, lineart, canny, segmentation and depth.}
    \label{tab:dataprompt}
    \begin{tabular}{p{.96\linewidth}}
    \toprule
    I'll provide an original prompt and two images. The first image is a \{condition\_type\} based control image for the conditional image generation task, and the second image is the ground truth image. I want you to analyze the \{condition\_type\} based control image and the original prompt, then expand it into a new prompt that can closely describe the ground truth image. First, output your reasoning process within $\mathrm{<think></think>}$ tags (your reasoning must not include analysis of the ground truth image). Then provide the expanded prompt within $\mathrm{<answer></answer>}$ tags. Original prompt: "\{original\_prompt\}" \\
    \bottomrule
    \end{tabular}
    \vspace{-5mm}
\end{table*}

\begin{table*}[t]
    \centering
    \caption{Template of the system prompt for the Two Phase Training. Condition types include hed, lineart, canny, segmentation and depth.}
    \label{tab:dataprompt2}
    \begin{tabular}{p{.96\linewidth}}
    \toprule
    You are a professional prompt engineer tasked with rewriting image-generation prompts based on a provided "\{condition\_type\} based control image" and an original prompt. The user will provide an original prompt and a \{condition\_type\} based control image which is for the conditional image generation. Your task is to analyze the \{condition\_type\} based control image and the original prompt, then expand it into a new prompt that can closely describe the image to generate. First, output your reasoning process within $\mathrm{<think></think>}$ tags. Then provide the expanded prompt within $\mathrm{<answer></answer>}$ tags. \\
    \bottomrule
    \end{tabular}
    \vspace{-5mm}
\end{table*}

\begin{table*}[t]
    \centering
    \caption{Template of the user prompt for the Two Phase Training. Condition types include hed, lineart, canny, segmentation and depth.}
    \label{tab:dataprompt3}
    \begin{tabular}{p{.96\linewidth}}
    \toprule
    The control image: $\mathrm{<image>}$. Original Prompt: \{original\_prompt\} \\
    Please analyze and rewrite the above image-generation prompt based on the provided \{condition\_type\} based control image and the original prompt. \\
    \bottomrule
    \end{tabular}
    \vspace{-5mm}
\end{table*}